\documentclass{article}
\usepackage{spconf,amsmath,graphicx}
\usepackage[usenames,dvipsnames]{xcolor}
\usepackage{soul}
\usepackage{enumitem}
\usepackage{comment}
\usepackage[hyphens]{url}

\usepackage{tabularx} 
\usepackage{booktabs} 

\title{Exploring the Impact of Moire Pattern on Deepfake Detectors}

\name{Razaib Tariq$^{\star}$ \qquad Shahroz Tariq$^{ \dagger}$ \qquad Simon S. Woo$^{\star}$}
  
  \address{$^{\star}$ Computer Science \& Engineering Department \\ Sungkyunkwan University, South Korea \\
      $^{\dagger}$ CSIRO's Data61, Australia}
      
\begin{document}
%
\maketitle
\begin{abstract}

Deepfake detection is critical in mitigating the societal threats posed by manipulated videos. While various algorithms have been developed for this purpose, challenges arise when detectors operate externally, such as on smartphones, when users take a photo of deepfake images and upload on the Internet. One significant challenge in such scenarios is the presence of Moiré patterns, which degrade image quality and confound conventional classification algorithms, including deep neural networks (DNNs). The impact of Moiré patterns remains largely unexplored for deepfake detectors. In this study, we investigate how camera-captured deepfake videos from digital screens affect detector performance. We conducted experiments using two prominent datasets, CelebDF and FF++, comparing the performance of four state-of-the-art detectors on camera-captured deepfake videos with introduced Moiré patterns. Our findings reveal a significant decline in detector accuracy, with none achieving above 68\% on average. This underscores the critical need to address Moiré pattern challenges in real-world deepfake detection scenarios.

\end{abstract}

\section{Introduction}
\label{sec:intro}

Deepfakes, synthetic media generated through advanced machine learning techniques, have emerged as a formidable and pressing concern in contemporary society. The proliferation of user-friendly, open-source deepfake generation tools exacerbates the problem by facilitating the dissemination of manipulated media. Amidst this landscape, developing robust detection mechanisms has become paramount to safeguarding the integrity of visual media and combating the potential harms posed by deepfakes~\cite{ShahrozAWS,WDC_Metaverse}.

Various algorithms have been proposed to address the detection of deepfake videos, typically leveraging frames extracted from the videos to discern their authenticity~\cite{SoK}. This approach proves effective in scenarios where the deepfake detector operates on the same device as the video being analyzed, facilitating internal processing. 

However, let us consider the scenario where malicious users take photos of deepfake videos and upload them to social networking sites. This can be considered as a new \textit{deepfake detector spoofing attack} to defeat the deepfake detector. In such a scenario which is possible and easily achievable, the detection paradigm undergoes a significant shift when the deepfake detector resides on an external device, such as a smartphone. At the same time, the target video exists on a separate device or platform. In such cases, the detection workflow necessitates the use of the smartphone camera to capture the deepfake content displayed on the screen of another device, thereby introducing complexities and challenges.

One challenge that commonly arises in this setting is the presence of Moiré patterns as shown in Figure 1, intricate visual artifacts that frequently manifest when capturing images or videos on digital screens.  These patterns pose a significant obstacle to accurate classification, as many existing algorithms, despite their reliance on deep neural networks (DNNs), struggle to effectively discern the presence of various patterns~\cite{goodfellow2015explaining}. Consequently, the failure to detect these artifacts often leads to erroneous classifications by the model, undermining the reliability of DNNs.

The primary objective of this paper is to empirically investigate the impact of camera-captured deepfake videos contaminated with Moiré patterns on the predictive performance of deepfake detectors. This is a new spoofing attack against the deepfake detectors. Therefore, in this work, specifically, we focus on evaluating the performance and possible degradation in detection accuracy when deepfake videos are captured through a smartphone camera and subsequently analyzed by state-of-the-art (SOTA) deepfake detection models~\cite{Cored,Xception,BZNet,QAD,ADD, CLRNet}. Our experimental analysis encompasses two widely-used deepfake datasets, namely CelebDF~\cite{li2019celeb} and Faceforensics (FF++)~\cite{rossler2019faceforensics++}, enabling a comprehensive assessment of detector performance across diverse deepfake scenarios.

\begin{figure*}[t]
    \centering
    \includegraphics[width=0.33\linewidth]{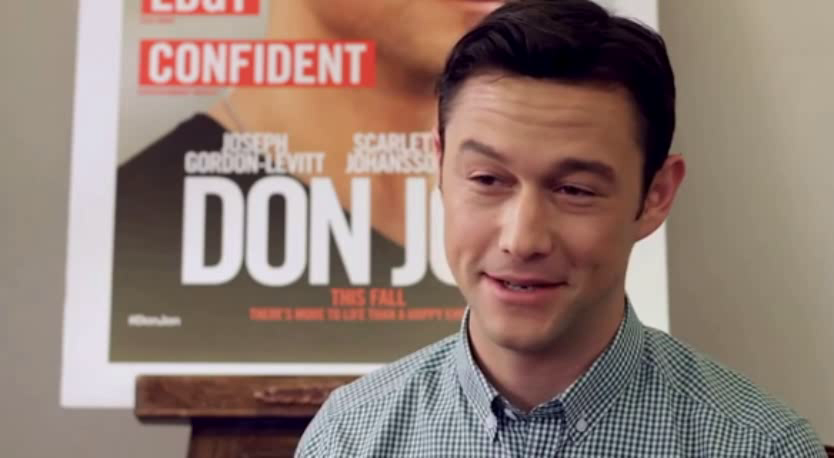}\hfill
    \includegraphics[trim={0pt 0pt 0pt 28pt},clip,width=0.33\linewidth]{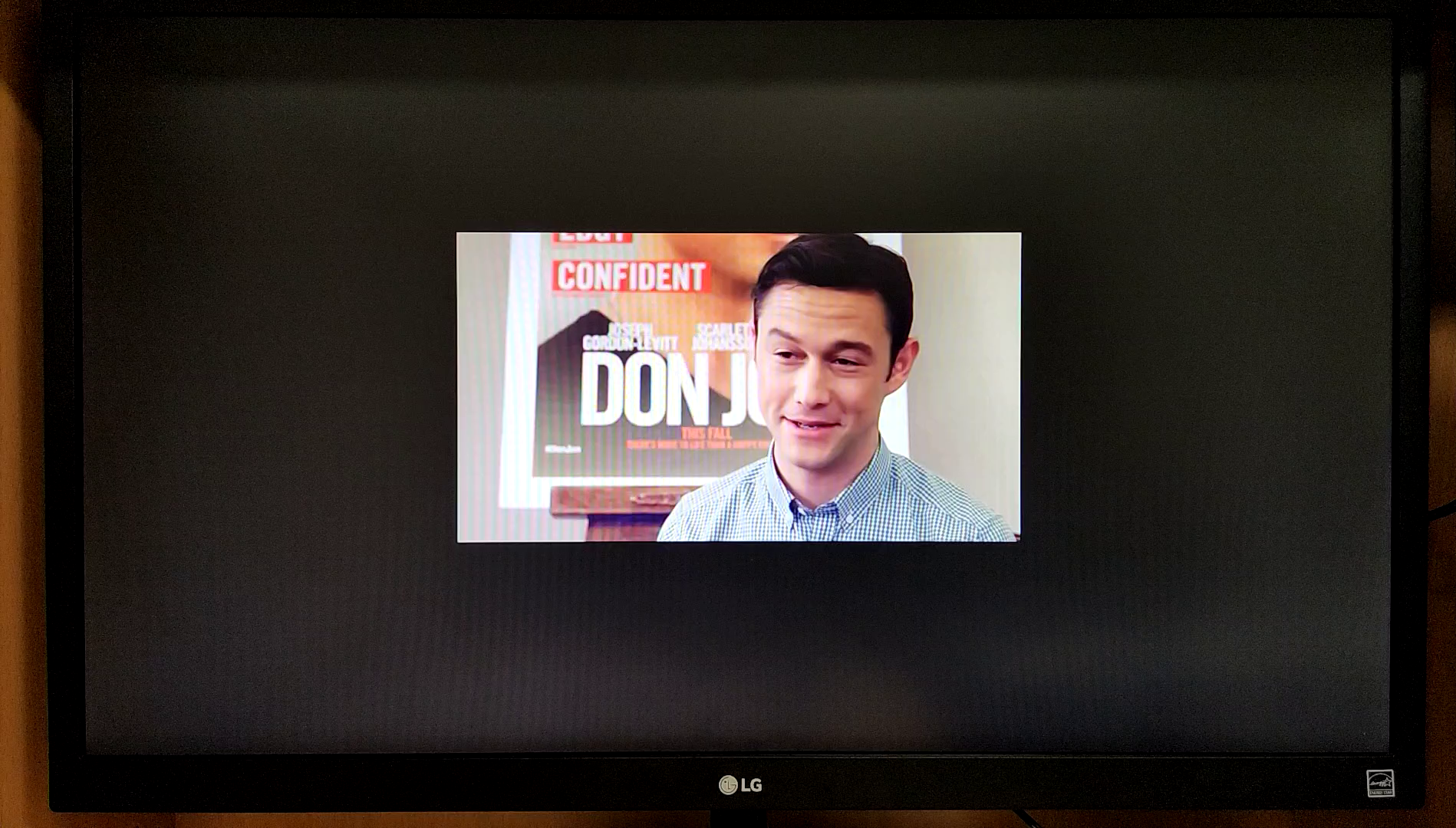}\hfill
    \includegraphics[width=0.33\linewidth]{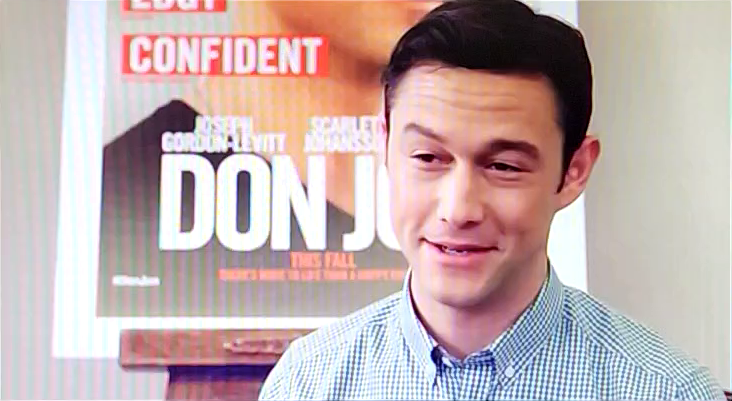}
    \includegraphics[width=0.33\linewidth]{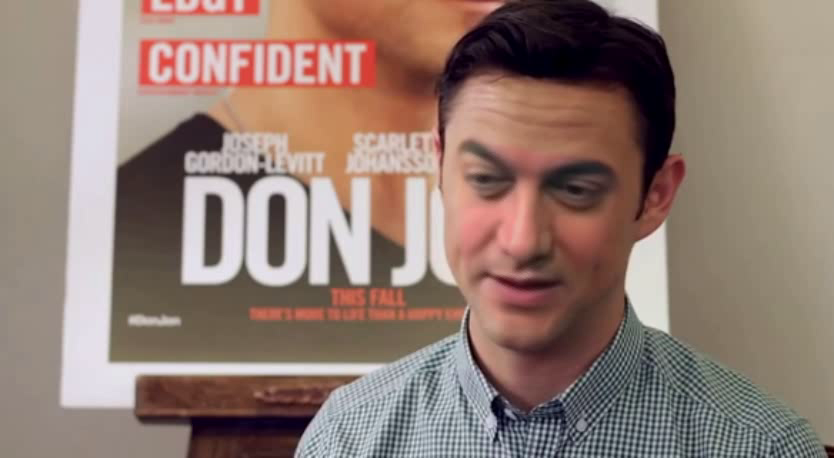}\hfill
    \includegraphics[trim={0pt 0pt 0pt 28pt},clip,width=0.33\linewidth]{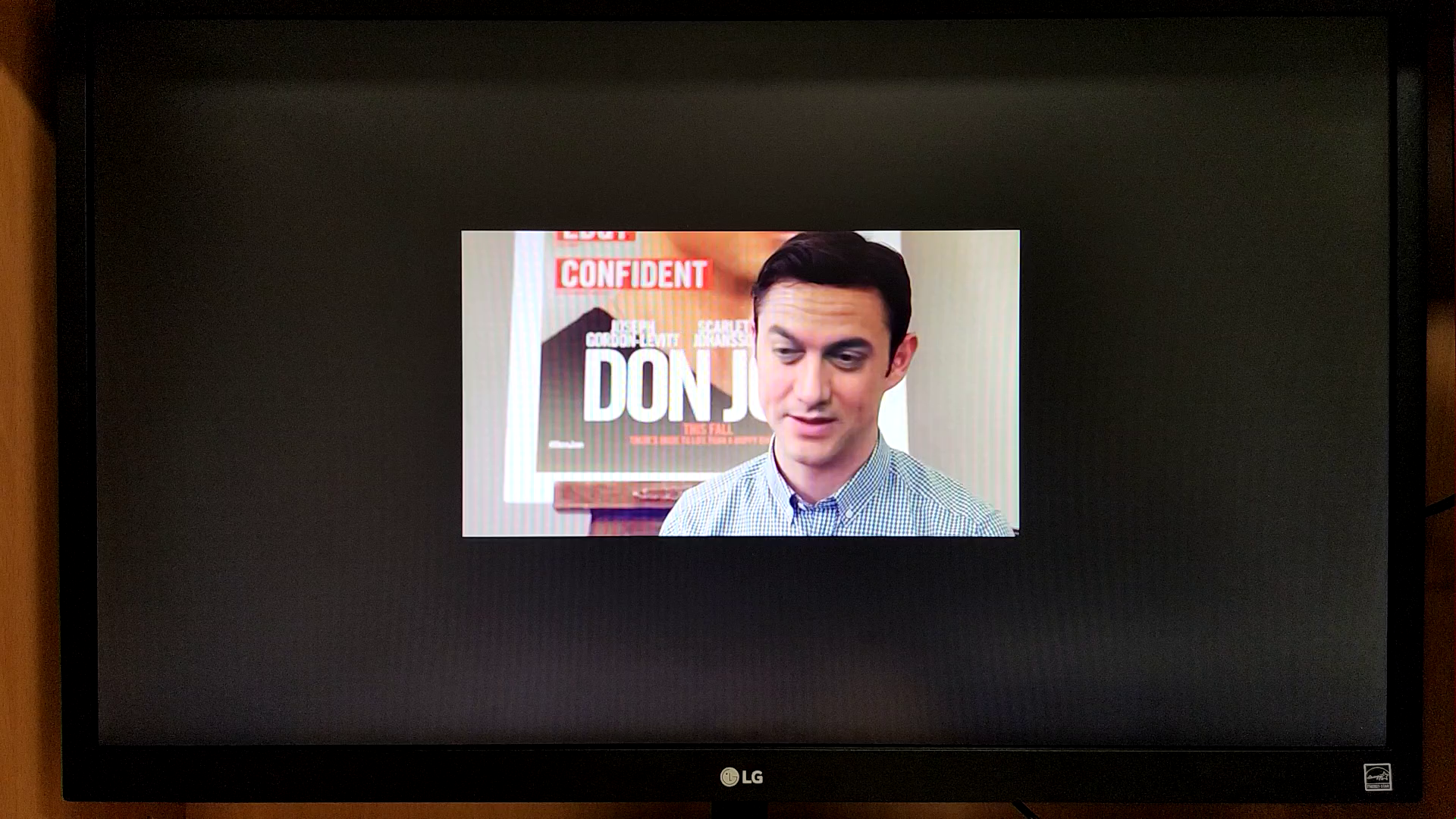}\hfill
    \includegraphics[width=0.33\linewidth]{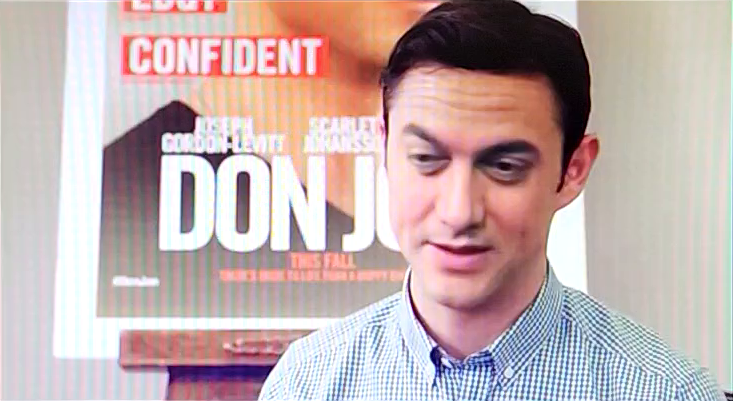}
    \caption{ Comparison between an original frame (without Moiré pattern) and a camera-captured frame (with Moiré pattern). The top row displays a frame extracted from an authentic video, while the bottom row presents a frame from a deepfake video. The left column depicts the original frame (without the Moiré pattern), the center column demonstrates its appearance on a computer screen, and the right column provides a magnified view, clearly revealing the presence of the Moiré pattern. }
    \label{fig:compare}
\end{figure*}

Through comparative analysis, we demonstrate the pronounced decline in detection efficacy when Moiré patterns are introduced into the analyzed videos, particularly in the context of camera-captured footage. By elucidating these findings, we aim to underscore the critical importance of accounting for Moiré patterns and similar artifacts in the development and deployment of robust deepfake detection systems. Ultimately, our research endeavors to contribute valuable insights toward enhancing the resilience of deepfake detection mechanisms in the face of evolving challenges. To the best of our knowledge, this is the first study to explore the impact of the Moiré pattern on deepfake video detectors. The main contributions of our work are as follows:

\begin{itemize}
    \item \textbf{Moiré Pattern's Impact on Deepfake Detection.} We conduct a systematic empirical study to assess the influence of camera-captured deepfake videos, particularly contaminated with Moiré patterns, on the predictive accuracy of state-of-the-art deepfake detectors.
    \item \textbf{Comprehensive Evaluation Across Diverse Datasets.} By leveraging two widely-used deepfake datasets, CelebDF and FF++, we provide a thorough evaluation of detection accuracy across diverse deepfake scenarios, offering insights into the generalizability and robustness of SOTA deepfake detectors against a new spoofing attack leveraging Moiré patterns. 
    \item \textbf{Identification of Vulnerabilities in SOTA Detectors.} Through comparative analysis, we identify significant declines in detection efficacy when Moiré patterns are introduced into camera-captured deepfake videos, highlighting vulnerabilities in state-of-the-art deepfake detection models and emphasizing the critical need for addressing artifact-induced challenges in deepfake detection systems.
\end{itemize}

The rest of the paper is organized as follows: In Section 2, we will discuss the related work on deepfake detection, as well as the challenges posed by the Moiré pattern. Next, in Section 3, we will provide our evaluation methodology, including the experimental settings. In Section 4, we will present the results of our evaluation and insights. In Section 5, we will provide a discussion about our findings, and finally, in Section 6, we will conclude the paper.

\section{Related Works}
\label{sec:related}

The rapid proliferation of deepfake technology has spurred a growing body of research aimed at developing robust detection mechanisms to mitigate the potential societal harms posed by synthetic media manipulation. In this section, we provide an overview of existing literature relevant to our investigation, focusing on prior works that address deepfake detection and the challenges posed by Moiré patterns in camera-captured footage.

\subsection{Deepfake Detection} Numerous studies have explored various methodologies for detecting synthetic media, leveraging a diverse array of features and techniques~\cite{MinhaFRETAL,ShallowNet1,SamTAR,JeonghoPTD,CLRNetold,HasamACMMM}. Traditional approaches often relied on handcrafted features or heuristic-based methods to discern anomalies indicative of synthetic manipulation~\cite{heuristic}. However, with the advent of deep learning, Convolutional Neural Networks (CNNs) have emerged as a dominant paradigm for deepfake detection~\cite{SoK,DeepfakeSurvey}. These models leverage the inherent spatial hierarchies within images and videos to learn discriminative features for distinguishing between genuine and manipulated content. 

\subsection{Moiré Patterns and DNNs} Despite the effectiveness of DNN-based approaches for various classification tasks~\cite{Russakovsky2014ImageNet}, challenges persist in accurate classification, particularly in scenarios involving camera-captured footage. One such challenge arises from the presence of Moiré patterns, which can significantly degrade the quality of images and videos captured from digital screens. The intricate nature of these patterns poses a formidable obstacle to existing detection algorithms, often leading to misclassifications and reduced detection accuracy. However, it has not been studied in the context of deepfake videos. In this work, we aim to fill this gap.


\begin{figure}[t!]
     \centering
     \frame{\includegraphics[width=0.6\linewidth]{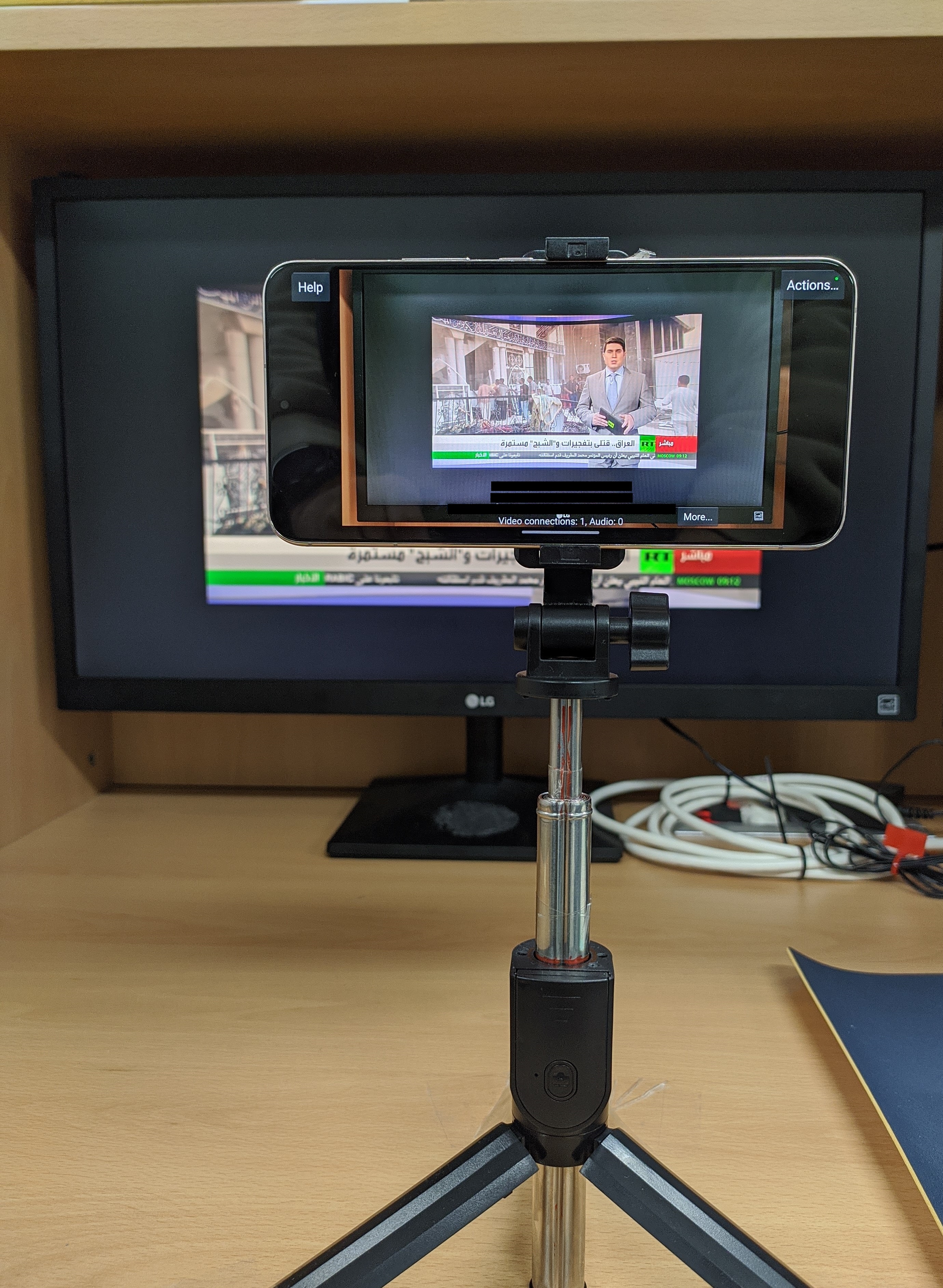}}
     \caption{The experimental setup involves the playback of videos sourced from FF++ and CelebDF on a computer screen. Subsequently, these videos are recorded using a smartphone camera. This method allows for the capture of the inherent Moiré pattern induced on the computer screen within the recorded video.}
     \label{fig:setup}
\end{figure}
\section{Evaluation Methodology}
\label{sec:method}

To evaluate the impact of Moiré patterns on the performance of state-of-the-art (SOTA) deepfake detectors, we conducted a comprehensive experimental study. Our evaluation methodology aimed to simulate real-world scenarios where deepfake detection occurs across different devices, particularly focusing on scenarios involving the use of smartphone cameras to capture deepfake content displayed on external screens.

\subsection{Datasets for Experiments}
We utilized two well-established deepfake datasets, CelebDF and Faceforensics++ (FF++), to ensure the diversity and representativeness of our data. CelebDF comprises videos featuring celebrities, while FF++ offers a broader range of synthetic and manipulated videos across various contexts. From the FF++ dataset, we selected the DeepFake, Face2Face, FaceSwap, and Neural Texture datasets.

\subsubsection{Videos Selection} We used a subset of videos from the FF++ and CelebDF Datasets for our experiment. To mitigate any potential biases in the video selection process, we employed a Python script to randomly retrieve videos from each dataset. We provide the total number of videos used in our experiment in Table 1.

\subsubsection{Moiré Pattern Introduction} For the purpose of introducing Moiré patterns into these deepfake videos, we started by displaying them on a computer screen and then recording them with the camera on our smartphone, as illustrated in our setup in Fig.~\ref{fig:setup}.

\begin{enumerate}
    \item \textbf{Equipment and Configuration:}  We employed a Samsung S22 Plus smartphone mounted on a tripod and an LG monitor with a resolution of 1980×1080. The setup allowed for the observation of Moiré patterns through the smartphone camera. To maintain consistency in face detection across videos, we fixed the tripod at a specific distance from the monitor.

    \item \textbf{Video Capturing Process:} We recorded the videos using the IP-webcam Android application, downloaded from the Google Play Store onto the Samsung S22 Plus smartphone. The default camera settings were used to replicate real-world scenarios. The monitor was positioned 30 centimeters away from the tripod at a height of 28 centimeters, with its resolution set to the default refresh rate of 1980×1080. Full-HD videos were captured at 30 frames per second with a bitrate of 16,000 kbit/s from the camera.

\end{enumerate}

\subsubsection{Camera-Captured Videos}
\label{sec:camera-captured_videos}
Through the process method above, we captured 200 deepfake videos from the FF++ dataset (100 real and 100 fake) equally divided among the four sub-datasets in FF++. In CelebDF, there are videos of 58 unique celebrities. Therefore, we captured one real and one fake for each celebrity, resulting in a total of 116 videos from CelebDF. Overall, the total videos used in the experiments are 316, each containing 350+ frames (i.e., approx. $316 \times 350=110,600$ video frame or images) shown in Table 1.




\begin{figure*}
    \centering
    \includegraphics[trim={30pt 25pt 50pt 18pt},clip, width=\linewidth]{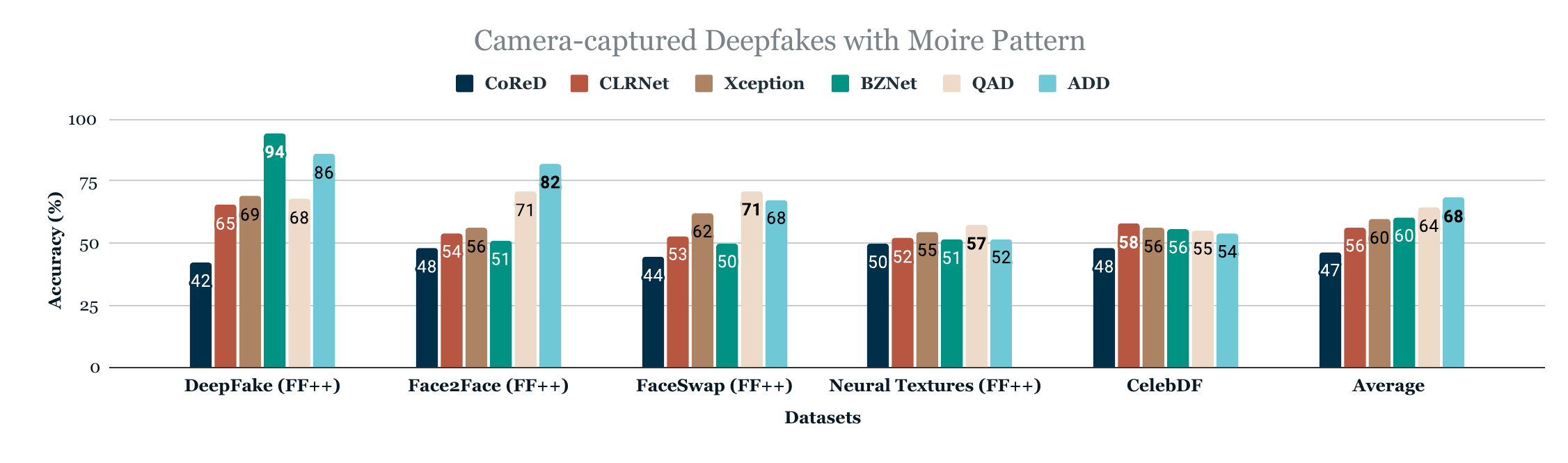}
    \caption{Average F1-score performance of five state-of-the-art (SOTA) deepfake detectors, when tested on camera-captured deepfake videos played on a computer screen, which introduce Moiré patterns. These videos are sourced from the FaceForensics++ and CelebDF datasets.}
    \label{fig:results}
\end{figure*}

\subsection{Experimental Setup}
We designed our experiments to compare the performance of deepfake detectors as follows:

\subsubsection{Experiment Procedure} After finishing the process of capturing the video. Following that, we proceeded to put these videos that were captured by the camera through the detectors in order to evaluate their effectiveness in the presence of Moiré pattern artifacts.

\begin{table}[t!]
\caption{Dataset Distribution.}
\centering
\small 
\begin{tabularx}{\columnwidth}{Xccc} 
\toprule
Datasets & Total Videos & CC-Real Videos & CC-Fake Videos \\ 
\hline
\midrule
CelebDF & 116 & 58 & 58 \\
DF~(FF++) & 50 & 25 & 25 \\
F2F~(FF++) & 50 & 25 & 25 \\
FS~(FF++) & 50 & 25 & 25 \\
NT~(FF++) & 50 & 25 & 25 \\
\bottomrule
\end{tabularx}
\label{table:resource}
\end{table}

\subsubsection{Deepfake Detectors}
We selected four state-of-the-art deepfake detectors representing a diverse range of methodologies and approaches, as follows: ADD~\cite{rossler2019faceforensics++}, Xception\cite{Xception}, CLRNet~\cite{CLRNet}, and QAD~\cite{QAD}. These detectors have demonstrated high performance and robustness across various benchmark datasets. The detectors were trained on clean deepfake datasets (i.e., w/o Moiré pattern) using the specifications provided in their original papers.

\subsubsection{Evaluation Metrics}
To quantify the impact of Moiré patterns on the deepfake detector's performance, we employed standard evaluation metrics of the F1-score. It is one of the most commonly used metrics for the performance assessment of deepfake detectors.

\section{Results}
\label{sec:results}Our findings from the experiments that we conducted are presented in~\ref{fig:results}. We discuss some of the most important findings and insights that came from the results.

\subsection{Performance Degradation of SOTA detectors.}
The SOTA deepfake detectors we evaluated, including CoReD, CLRNet, Xception, BZNet, QAD, and ADD, typically exhibit accuracy rates in the high 90s when tested on benchmarking datasets like FF++ and CelebDF. However, we have observed a notable performance degradation when these detectors encounter deepfake videos incorporating Moiré patterns. In such cases, performance metrics plummet to as low as 47\% for CoRed, with even the highest-performing detector, ADD, unable to surpass 68\%. This issue raises significant concerns, as naturally occurring patterns introduced during real-world application have the potential to undermine the credibility of these detectors by yielding erroneous predictions.

\subsection{Interesting Insights and Findings}
\begin{enumerate}
    \item \textbf{Finding \#1.} The performance of BZNet on the DeepFake (FF++) dataset demonstrates promise at 94\%; however, it exhibits a near-random guess level (i.e., approximately 50\%) on all other datasets, indicating potential overfitting to the DeepFake (FF++) dataset. Consequently, the overall performance of BZNet across the five datasets is 60\%, as illustrated in the `Average' section of Fig.~\ref{fig:results}.
    \item \textbf{Finding \#2.} Like BZNet, ADD shows relatively high performance at 86\%. However, unlike BZNet, ADD also performs well on the Face2Face (FF++) dataset, suggesting a higher degree of generalizability and positioning it as the best-performing detector among the five evaluated in our experiment, with 68\% on average. Nonetheless, it's important to note that this performance falls short of the typical high 90s range expected from these deepfake detectors. Therefore, the detectors need to be equipped with techniques to identify and mitigate Moiré patterns during training to achieve acceptable real-world performance levels (i.e., in the high 90s).
\end{enumerate}

\begin{figure}
    \centering
\includegraphics[trim={10pt 0 500pt 0},clip,width=1\linewidth]{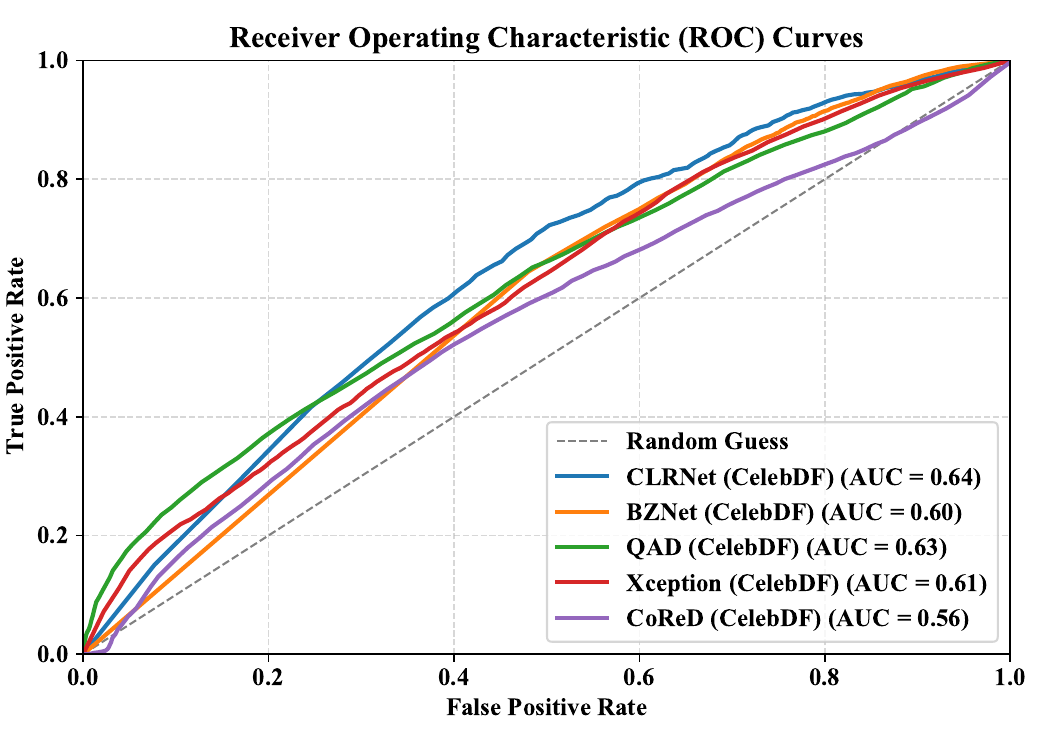}
\includegraphics[trim={512 0 0pt 0},clip,width=1\linewidth]{figures/CelebDF_CC.pdf}
    \caption{The AUROC (top) and Precision-Recall (bottom) curves of deepfake detectors against camera-captured deepfakes (containing Moiré patterns) from the CelebDF dataset.}
    \label{fig:AUC_PR}
\end{figure}

\subsection{AUROC Scores}
We present the AUROC scores of all evaluated detectors on the CelebDF dataset in Fig.~\ref{fig:AUC_PR}. The AUROC scores align closely with the discussed results. Notably, CLRNet achieves the highest AUROC score of 0.64, which is notably low. This indicates limited discriminative power against camera-captured deepfakes with Moiré patterns, signaling the need for substantial improvements. The relatively low AUROC scores across the evaluated detectors underscore the challenges in effectively detecting deepfake videos, particularly those featuring Moiré patterns. Further research and refinement of detection techniques are imperative to enhance the robustness and efficacy of deepfake detection systems.

\subsection{Precison-Recall (PR) Curves}
The PR curves illustrate the trade-off between precision and recall across varying threshold values for the deepfake detection model. As we can observe, the curves start at high precision (except for CoReD), indicating strong performance at higher threshold values where the model is conservative in making positive predictions. However, as the threshold decreases, allowing the model to be less conservative, recall increases while precision decreases. This behavior is characteristic of models where adjusting the threshold affects the precision-recall balance. Notably, the PR curves reveal dips below the no-skill line for the CoReD model, indicating varying performance across different threshold values. These fluctuations highlight the sensitivity of the model's performance to threshold adjustments and underscore the importance of fine-tuning parameters to optimize detection performance. As depicted in Fig.~\ref{fig:AUC_PR}, understanding the intricacies of precision-recall trade-offs is crucial for effectively evaluating and improving deepfake detection systems.

\section{Discussion}
\label{sec:discussion}

\subsection{Incorporation of Novel Preprocessing Techniques}
The results of our experiments demonstrated that Moiré patterns have a significant influence on the accuracy to which deepfake detectors present their results. As a consequence of this, there is an urgent requirement to develop innovative approaches and procedures for these detectors that are capable of effectively mitigating the impact of Moiré artifacts. Because of this, it is necessary to incorporate defenses against Moiré patterns into the training methodology of the detectors, particularly during the stage of data preprocessing and through methods such as data augmentation. These kinds of improvements are absolutely necessary in order to improve the effectiveness of the detectors across a wider range of situations that occur in the real world.

\subsection{Dataset with Moiré Patterns}
While preprocessing and data augmentation techniques have shown some promise in enhancing deepfake detectors against Moiré-patterned deepfakes through the synthetic introduction of such patterns into the training data, it is important to note that this approach has inherent limitations in terms of its ability to replicate all environmental factors that are present in the real world. A comprehensive deepfake dataset that authentically captures these patterns in real-world settings using physical devices such as computer screens and smartphone cameras is therefore required.

In this paper, we present a preliminary version of such a dataset developed on a smaller scale for experimental purposes. Our future endeavor, on the other hand, intends to significantly expand this dataset by incorporating well-established deepfake benchmark datasets and introducing a wide range of environmental conditions. These conditions will include variations in camera angles, lighting conditions, screen types, and camera models. We argue that such a dataset has the potential to be the best solution for strengthening deepfake detectors against patterns that are induced by physical environmental factors, which would result in an increase in the robustness of these detectors in real-world deployment scenarios.

\subsection{Evaluation of Transferability Across Domains}
In advancing deepfake detection systems, a critical consideration lies in evaluating their transferability across various domains, including those influenced by recent advancements in generative AI and diffusion-based deepfakes. While a detector may effectively identify deepfakes generated using specific techniques or originating from certain sources, its performance may degrade when confronted with unseen or adversarially crafted deepfakes, as evidenced by our findings. Therefore, assessing detectors' robustness and generalization capabilities across diverse datasets and scenarios is essential, as emphasized in prior research~\cite{SoK}. Future endeavors should prioritize the development of evaluation methodologies that rigorously assess the transferability of deepfake detection models, thereby ensuring their reliability and suitability for real-world applications.

\subsection{Limitations and Future Research Prospects}
One of the most notable drawbacks of our paper is that it does not contain a more extensive selection of image and video detectors. These detectors are necessary for detecting deepfakes when Moiré patterns are present. Furthermore, our dataset was obtained by using a limited selection of smartphone devices, which may not accurately depict the wide variety of environmental variables that can cause Moiré to occur. In the future, research should incorporate a wider variety of image and video detectors in order to improve the accuracy of deepfake detection when it is confronted with Moiré patterns. It is absolutely necessary to collect datasets using a wide variety of smartphone devices in order to guarantee the reliability of the data acquired from a variety of monitor variations. Additionally, the incorporation of a variety of monitors that are capable of introducing distinct Moiré patterns when captured by smartphone cameras would provide a more comprehensive understanding of the impact that these monitors have on the detection of deepfakes. It is of crucial importance to develop methods that involve the reduction of Moiré patterns, as this will allow for the accurate detection of deepfake features. In order to determine the most effective strategies for evaluating deepfake detection methods, additional experimentation using these methods, in conjunction with a variety of detection techniques, will be invaluable value.

\section{Conclusion}
\label{sec:conclusion}

To the best of our knowledge, so far all the deepfake detectors have trained on non-Moiré patterned images or videos. In this work, we take into consideration a new type of spoofing attack that can be used against deepfake detectors by utilizing Moiré patterns. It is true that Moiré patterns frequently show up and appear convincing in the recording of computer screens that have been captured by cameras. Furthermore, malicious users have the ability to upload Moiré-added deepfakes to bypass the deepfake detectors.
In this study, we investigated the impact of Moiré patterns on deepfake detectors, revealing their susceptibility to such patterns, leading to considerable performance degradation. In addition, we discussed potential solutions to this problem, such as preprocessing techniques and the development of new datasets, in order to find a solution. Future research endeavors should prioritize addressing the challenges that have been mentioned above, while also working toward the development of detectors that are resistant to a variety of patterns that are artificially or naturally induced in videos. These patterns include Moiré patterns and adversarial noise types.

\section*{Acknowledgments} We thank the anonymous reviewers for their helpful comments. Simon S. Woo is the corresponding author. This work was partly supported by Institute for Information \& communication Technology Planning \& evaluation (IITP) grants funded by the Korean government MSIT: (RS-2022-II221199, Graduate School of Convergence Security at Sungkyunkwan University), (No. RS-2024-00337703, Development of satellite security vulnerability detection techniques using AI and specification-based automation tools), (No. 2022-0-01045, Self-directed Multi-Modal Intelligence for solving unknown, open domain problems), (No. RS-2022-II220688, AI Platform to Fully Adapt and Reflect Privacy-Policy Changes), (No. RS-2021-II212068, Artificial Intelligence Innovation Hub), (No. 2019-0-00421, AI Graduate School Support Program at Sungkyunkwan University), (No. RS-2023-00230337, Advanced and Proactive AI Platform Research and Development Against Malicious Deepfakes), and (No. RS-2024-00356293, AI-Generated Fake Multimedia Detection, Erasing, and Machine Unlearning Research). 


\bibliographystyle{IEEEbib}
\bibliography{references}

\end{document}